\documentclass[a4paper]{article}
\usepackage{appendix}
\usepackage{url,hyperref,lineno,microtype,subcaption}
\usepackage{mathtools, amsmath, amsfonts}
\usepackage[onehalfspacing]{setspace}
\usepackage{datetime}
\usepackage{fmtcount}
\usepackage{etoolbox}
\usepackage{fcprefix}
\usepackage{natbib}
\usepackage{booktabs}
\usepackage[a4paper, total={6in, 8in}]{geometry}
\usepackage{soul}
\usepackage{authblk}
\usepackage{xcolor}
\usepackage{placeins}
\usepackage{parskip}

\providecommand{\keywords}[1]{\textbf{\textit{Keywords ---}} #1}
\begin{document}
\onecolumn

\title{\textbf{Dynamic Time Warping as a New Evaluation for Dst Forecast with Machine Learning}}
\author{Brecht Laperre}
\author{Jorge Amaya}
\author{Giovanni Lapenta}
\affil[1]{Centre for mathematical Plasma Astrophysics, Department of Mathematics, KU Leuven, Leuven, Belgium}
\date{}

\maketitle

\begin{abstract}
Models based on neural networks and machine learning are seeing a rise in popularity in space physics. In particular, the forecasting of geomagnetic indices with neural network models is becoming a popular field of study. These models are evaluated with metrics such as the root-mean-square error (RMSE) and Pearson correlation coefficient. However, these classical metrics sometimes fail to capture crucial behavior. To show where the classical metrics are lacking, we trained a neural network, using a long short-term memory network, to make a forecast of the disturbance storm time index at origin time $t$ with a forecasting horizon of 1 up to 6 hours, trained on OMNIWeb data. Inspection of the model's results with the correlation coefficient and RMSE indicated a performance comparable to the latest publications. However, visual inspection showed that the predictions made by the neural network were behaving similarly to the persistence model. In this work, a new method is proposed to measure whether two time series are shifted in time with respect to each other, such as the persistence model output versus the observation. The new measure, based on Dynamical Time Warping, is capable of identifying results made by the persistence model and shows promising results in confirming the visual observations of the neural network's output. Finally, different methodologies for training the neural network are explored in order to remove the persistence behavior from the results. \footnote{Code: \url{https://github.com/brechtlaperre/DTW_measure}}
\end{abstract}

\keywords{Space Weather, Forecast, Dst, Machine Learning, DTW, Evaluation, LSTM, Storm}

\section{Introduction}
\label{sect:introduction}
The disturbance storm time (Dst) index is calculated from the measurements of four ground-based magnetometer stations, located close to the equator and spread evenly across the Earth \citep{sugiura1991equatorial}. The index, introduced by \citet{sugiura1963hourly}, is the average of the magnetic disturbance of the Earth's magnetic field horizontal component. Most often, the Dst is used as a measure of the strength of the axi-symmetric magnetosphere currents, capturing the dynamics of the inner magnetospheric current system. The most important types of these dynamics are geomagnetic storms. The Dst index can be used to identify the three phases of these storms: the initial phase, the main phase and the recovery phase. 

Geomagnetic storms are large perturbations in the Earth's magnetic field. They are caused by the coupling between solar wind and magnetosphere, in particular the southward component of the interplanetary magnetic field (IMF) \citep{burton1975empirical}. When magnetic reconnection happens between the IMF and the Earth's magnetosphere, an influx of energetic particles from the solar wind into the magnetosphere occurs. This increases the intensity of the Earth's ring current, which is reflected by the Dst index \citep{akasofu1981energy, gonzalez1994geomagnetic}. In the case of very intense geomagnetic storms, satellites and power grids can be damaged, as well as extensive pipeline systems such as those used to transport gas, oil and water \citep{kasinskii2007effect}. The most prominent examples of intense geomagnetic storms are the Carrington event and the 2003 Halloween storm. From these historical storms, it is clear that there will be global economical losses and damages when another intense geomagnetic storm would hit Earth \citep{NAP12507, kappernman1990bracing, pulkkinen2005geomagnetic}. Consequently, the ability to timely forecast geomagnetic storms has been a topic of interest in geophysics for the past three decades. 

Because of how the Dst index is linked to geomagnetic storms, forecasting the Dst is the most direct way of forecasting geomagnetic storms. \citet{burton1975empirical} was one of the first to construct a model of the Dst. Their linear data-driven model consisted of a driver and decay term. These terms were estimated using the available data of the solar wind velocity and density, and the southward component of the IMF. Others have tested and improved the initial model by modifying the driver and decay term \citep[see e.g.][]{Klimas1998, Temerin2002, temerin2006dst}, with \citet{Ji2012} providing an overview of these models. However, a linear coupling between solar wind and magnetosphere is inadequate for predicting large geomagnetic disturbances, and non-linear systems are required to fully capture the behavior of the Dst \citep{iyemori1990storm}. 

A popular approach to modeling this non-linearity is through neural networks. Feed-forward neural networks were used first for this application. \citet{Lundstedt1994} was one of the first to use this type of neural network, using as input the $B_z$ component of the IMF and the velocity and density of the solar wind to forecast the Dst 1h in advance. Later, \citet{Stepanova2000} used the previous Dst values as input to predict the Dst 3h in advance. \citet{Bala2012} was able to forecast the Dst 6h in advance by using the Boyle index as their most important input \citep{boyle1997empirical}. \citet{Lazzus2017} used a feed-forward neural network, but used particle-swarm optimization instead of backpropagation \citep{Eberhart1995} to train their model. Using the past Dst as an input, they were able to provide a forecast of the Dst up to 6h in advance, showing the benefit of this type of training algorithm. 

With the introduction of the recurrent neural network, in particular the Elman recurrent network \citep{elman1990finding}, new forecast models where introduced. \citet{wu1997geomagnetic} used an Elman recurrent network  to provide a forecast of the Dst index up to 6h in advance. Similarly, many more used an Elman recurrent network and input from the solar wind to forecast the Dst index \cite[see e.g.][]{Barkhatov2001, lundstedt2002operational, Pallocchia2006, watanabe2002prediction}. More recently, \citet{Gruet2018} used a long short-term memory (LSTM) neural network instead, and combined it with Gaussian processes to provide both a regressive and a probabilistic forecast of the Dst for 1h to 6h in advance \citep{hochreiter1997long}. 

However, some authors detected problems with the forecast of the Dst. \citet{Stepanova2000} used a feed-forward neural network and previous Dst values to provide a forecast up to 3h in advance. More advanced forecasts had shown a time shift between the observed and predicted Dst, forecasting geomagnetic storms too late. This effect was also detected by \citet{wintoft2018evaluation}, who evaluated forecasts of the Kp and Dst with their neural network up to 3h in advance. Their Kp forecast had a time shift between forecast and observation for 2h and 3h in advance, while Dst forecast of the main phase of geomagnetic storms showed time shifts at already 1h in advance. 

This paper aims to highlight this time shift problem. Section \ref{sect:Experiment} sets up an experiment where a recurrent neural network is trained to forecast the Dst 6h in advance. It compares our model with those in the literature, and concludes by highlighting the time shift observed in geomagnetic storm forecasts. Then, in Section \ref{sect:DTW} a new measure is introduced that is capable of accurately measuring this time shift between observation and prediction. Finally, Section \ref{sect:Discussion} looks further into why this time shift behavior is observed, and potential solutions.

\section[Experiment]{The experiment}
\label{sect:Experiment}
This section concerns the details of the experiment, together with the discussion of the results that lead to the discovery of the problems. The initial problem was to train a neural network to forecast the values of the Dst-index at times $t+l$, $l \ge 1$, while having information up to time $t$.

The layout is the following: first the data used to train the neural network model is explained. Second, the processing of the data is discussed. Then the method of evaluation is described. Afterwards, the neural network model is described in detail. Finally, the results of the model are analyzed and discussed.

The following terminology will be used throughout this section. \citet{box2015time} defines Dst($t+l$) as a forecast at \textit{origin} $t$ with a \textit{lead time} $l$. We will differ from this terminology, instead naming the Dst($t+l$) a forecast at \textit{forecasting horizon} $t+l$.

\subsection[Data]{The data}
\label{sect:exp:data}
The data used to train and test the neural network model were obtained from the NASA/GSFC's OMNI database \citep{king2005solar}. From the database, the hourly averages of the solar wind velocity $V_{sw}$ and density $\rho_{sw}$, the IMF z-component $B_z$ and magnitude $|B|$, together with the geomagnetic Dst index were extracted. These physical quantities will be referred to as `features' throughout the paper. In particular, these features were measured between 00:00, 14 January 2001 to 23:00, 31 December 2016, and the full extracted data set contains a total of 139,944 entries.

\subsection[Preprocessing]{Preprocessing the data}
\label{sect:exp:preprocessing}
The preprocessing of the data is done with the following steps. The data are first split into a training, test and validation set, to prevent information bias through validation leakage. Next, each set is scaled and normalized using scaling parameters measured from the training set. Finally invalid measurements are removed from each of the sets by using a sliding window. Each step of this procedure is discussed in more details below. 

Before assigning entries to one of the data sets, the data are split into monthly samples. The first sample corresponds to the month January of 2001, the final sample to the month December of 2016, equating to 181 samples. The reason for this split is the high temporal correlation of the data. Hourly samples are highly correlated, which causes the model to artificially perform better on the test set \citep{camporeale2019challenge}.

The data sets are constructed as follows. The test set consists of the months April, August and December of each year, corresponding to 25\% of the total data. These months have been chosen arbitrarily, and mainly ensure a good spread of the test data over the given time period. We expect to see the same kind of results when the test set would be taken differently. An experiment to determine the effect of the chosen test set on the performance will be done in Section \ref{sect:Experiment}. From the remaining data, 60\% of the months are distributed randomly into the training set. The remaining 15\% of the data are placed in the validation set. This corresponds to 77,544, 20,520, and 33,120 entries in training, validation and test set, respectively.

The choice of training, validation and test set plays a huge role in the performance of the model. A difference in these sets makes direct comparison of forecasting models difficult, as stated by \citet{Lazzus2017}. In order to measure the variance caused by this choice, 10-fold cross validation is performed and the results are reported in Section \ref{sect:exp:Results}.

After this step, the sets are scaled and normalized. This ensures that every feature lies within the same range of values, ensuring comparability of the different features and faster convergence of the machine learning algorithm \citep[see e.g.][]{juszczak2002feature}. The full transformation process is done in two steps: first the scaling constants are determined from the training set, then the transformation is applied on the training, validation, and test set. The features are transformed by removing the mean and scaling to unit variance:
\begin{equation} 
	\bar{X}_{train} = \frac{X_{train} - \mu_{train}}{\sigma_{train}}, \qquad \bar{X}_{valid} = \frac{X_{valid} - \mu_{train} }{\sigma_{train}},\qquad \bar{X}_{test} = \frac{X_{test} - \mu_{train} }{\sigma_{train}},\label{eq:normalization}
\end{equation}
where $X$ is the data set, $\mu_{train}$ the mean of the training data set and $\sigma_{train}$ the standard deviation of the training set.

The final step of the preprocessing extracts valid time series window samples from each set. A sample is valid if, for every feature, there are no missing or invalid measurements. The output can be ignored, because the Dst index contains no missing values. The samples are extracted by moving a sliding window over the full data set. The size of the sliding window is equal to the size of the length of the time series used as input for the model, which in our case will have a length 6 hours. We chose 6 hours based on results of \citet{Lazzus2017}, who evaluated their time series input from length $t$ back to $t-48h$ through an exhaustive procedure, and found no significant improvement in the forecast when using data from times further in the past than $t-6h$. The target of the forecast is the Dst index from times $t+1h$ to $t+6h$. We chose $6h$ to be able to compare to other results found in the literature. After the preprocessing, there are 74,117, 19,596, and 32,166 valid samples in the training, validation, and test set, respectively. This corresponds to about 96\% of the initial total data.

\subsection{Evaluation of the model}
\label{sect:exp:eval}
Throughout the paper, evaluation of the model's forecast is done through three distinct methods. The first method compares the neural network model to a naive forecasting model, a so-called baseline model. The second method is evaluation by use of a set of metrics. The third method is k-fold cross-validation (k-fold CV), a statistical method to test the model's predictive capabilities. Al of these methods are explained in this section.
\hphantom{ }

\subsubsection{Baseline model}
\label{sect:exp:eval:baseline}
The baseline model is a simplified empirical law that can generate a zero order forecast. The most simple type of time-series forecast is done with the persistence model. This model assumes that the value at the current time step does not change, so the next time step is predicted as:
\begin{equation}
	Dst(t+1h) = Dst(t).
\end{equation} 
This model is easily extended to forecast multiple hours in the future:
\begin{equation}
	Dst(t+p) = Dst(t), \quad p \in \mathbb{N}.
\end{equation}
The work by \citet{owens201327} has shown that the persistence model can be a reliable predictor for some solar wind parameters, comparable to numerical models when evaluated in a point-by-point assessment method. In particular, geomagnetic activity and solar wind speed show good results when evaluated with a 27-day persistence model, and can be used as a benchmark for more sophisticated models.

\subsubsection{Metrics}
\label{sect:exp:eval:metrics}
Now the set of metrics used for the model evaluation are defined. The root mean square error (RMSE) and the Pearson linear correlation coefficient (R) are often used for the evaluation of time series. In addition, the set of fit performance metrics recommended by \citet{liemohn2018model} will also be used. These are the linear fit parameters, the mean absolute error (MAE), the mean error (ME) and the prediction efficiency (PE). All these metrics and their definitions are summed up in this section. Define $M_i$ as the forecast of the model and $O_i$ the corresponding real observational value, with $i = 1, \dots, N$, and $N$ the number of samples.
\begin{itemize}
\item The RMSE is defined as:
\begin{equation} \label{eq:eval:RMSE}
	RMSE = \sqrt{\frac{1}{N} \sum_{i=1}^N (M_i - O_i)^2}.
\end{equation}
This metric puts emphasis on outliers, which in the case of the Dst index corresponds to geomagnetic storms. A low RMSE thus corresponds to good accuracy of the forecast of geomagnetic storms.
\item The Pearson linear correlation coefficient (R), is given by:
\begin{equation} \label{eq:eval:R}
	R = \frac{\text{cov}(M_i, O_i)}{\sqrt{\text{var}(M_i) \text{var}(O_i)}}.
\end{equation}
This correlation coefficient gives a more global view of the prediction. It indicates if the model predicts the trend of the index correctly.
\item Since the model $M$ is predicting the observations $O$, a linear relationship is expected. Computing the linear fit of the model $M$ to the observations $O$ allows the results to be evaluated independent of time. Call $A$ the offset of the linear fit, and $B$ the slope. Then these parameters linearly relate $M$ to $O$ as follows:
\begin{equation} \label{eq:eval:LinearFit}
	M_i = A + B \cdot O_i.
\end{equation}
In the case of a perfect prediction, $B$ is 1 and $A$ is 0.
\item The MAE emphasizes the ``usual'' state of the index, which in the case of the Dst index corresponds to quiet time, where no geomagnetic storms are happening. The MAE indicates how well the model predicts these values. This value is defined as
\begin{equation}
	\label{eq:eval:MAE}
	MAE = \frac{1}{N} \sum_i^N |M_i - O_i|.
\end{equation}

\item The ME indicates if the model systematically over- or under-predicts the observations, based on its  positive or negative sign. If the mean error is zero, under and over predictions are balanced:
\begin{equation}
	\label{eq:eval:ME}
	ME = \frac{1}{N} \sum_i^N (M_i - O_i).
\end{equation}

\item Finally, the PE is used to quantify the model's ability at reproducing the time variation of the Dst index:
\begin{equation}
	\label{eq:eval:PE}
	PE = 1 - \frac{\sum_{i=1}^N (M_i - O_i)^2 }{\sum_{i=1}^N (O_i - \bar{O})^2 },
\end{equation}
where $\bar{O}$ is the average of the observational values. The maximum value of the PE is 1, corresponding to a perfect prediction at all times. A prediction efficiency equal to or less than zero indicates that the model is incapable of forecasting the time variation seen in the observations. 
\end{itemize}
\hphantom{ }

\subsection{K-fold cross-validation}
K-fold CV is a training technique for the model in order to learn its capabilities, without having to make use of the test set and so prevent an information bias. When applying CV, the training and validation sets defined in Subsection \ref{sect:exp:data} are used, unless stated otherwise. 

The technique firsts randomly splits the (monthly) data into $k$ equal-sized partitions (or folds). One of these folds is picked as the validation set, and the other remaining folds are used as the training set. The model is trained on the training set and evaluated on the validation set, and the evaluation is stored. This is repeated until every fold has been used exactly once as the validation set. By taking the average of the result of each run, an estimate of the predictive performance is given. This allows us to evaluate the model on different parameters, both hyperparameters and features, while preventing any optimization on the test set, as this would otherwise invalidate our results.

\hphantom{ }
\subsection{The model}
\label{sect:exp:model}
A neural network model is constructed and trained to forecast the Dst index for forecasting horizons $t+1h$ up to $t+6h$. As input, the model receives a multidimensional time series $X_t$, containing the data described in Subsection \ref{sect:exp:data}, ranging from time $t-6h$ to time $t$, as displayed in Equation \eqref{eq:InputRep}. The output then consists of a 6-dimensional vector $Y_t$, corresponding to the forecast Dst values for forecasting horizon $t+1h$ up to $t+6h$. 
\begin{equation} \label{eq:InputRep}
	\underbrace{
		\begin{pmatrix}
			V_{sw}(t-6h) & V_{sw}(t-5h) & \dots & V_{sw}(t)\\
			|B|(t-6h) & |B|(t-5h) & \dots &  |B|(t) \\
			\vdots & \vdots & \ddots & \vdots \\ 
			Dst(t-6h) & Dst(t-5h) & \dots & Dst(t)
		\end{pmatrix}
	}_{X_t} \rightarrow
	\underbrace{
		\begin{pmatrix}
			Dst(t+1h) \\ Dst(t+2h) \\ \vdots \\ Dst(t+6h)
	\end{pmatrix}}_{Y_t}
\end{equation}

The programming language Python, and in particular the package PyTorch \citep{paszke2017pytorch}, was used to implement and train the neural network. A link to the source code can be found in Appendix \ref{appendix:additionalInfo}.

\hphantom{ }
\subsubsection{Model description and training}
\begin{figure}[ht]
	\centering
	\includegraphics[width=\textwidth]{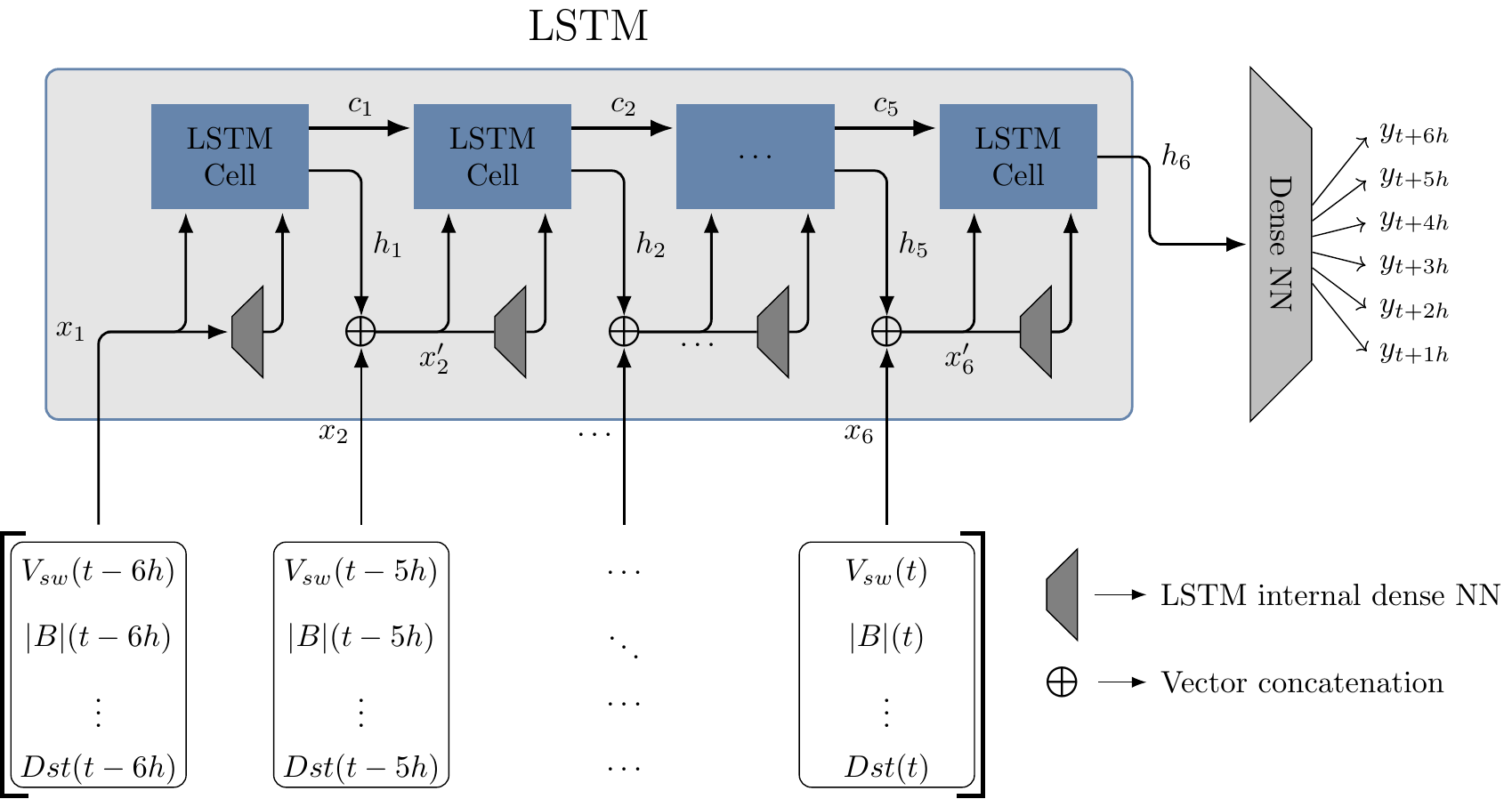}%
	\caption{The architecture of the LSTM-NN model. The time series input is first given to an LSTM, which iteratively uses each time step of the input $x_t$, together with the previous hidden state $h_{t-1}$, to update the memory state $c_t$ and construct a new hidden state $h_t$. The LSTM cell requires as input the concatenated vector from $x_t$ and $h_{t-1}$, together with the output of the LSTMs internal dense NN taking that same vector as input. The final hidden state of the LSTM is given to a dense neural network, who transforms it to a six-dimensional output vector $Y_t = \{y_{t+1h}, y_{t+2h}, \dots, y_{t+6h}\}$. The output $Y_t$ represents the forecast Dst, as displayed in Equation \eqref{eq:InputRep}.} \label{fig:model:model}
\end{figure}

The model architecture of the neural network model consists of a LSTM, combined with a dense neural network, as shown in Figure \ref{fig:model:model}. 

LSTMs are a type of recurrent neural network, where the output of the previous iteration is used as an additional input. When multiple past events are used as inputs, classic recurrent neural networks loose training information in a phenomena called gradient vanishing \citep{hochreiter1998vanishing}. LSTMs are designed to work better for long time series, by incorporating two internal memory states: the hidden state, $h_t$, and the memory state, $c_t$. A detailed explanation of how these memory states retain information and how they work is presented in \citet{hochreiter1997long}. We refer the reader to this publication for mode details on the exact internal workings of the LSTM.

The LSTM memory machinery is encapsulated in the LSTM cell. Figure \ref{fig:model:model} shows how the input, $X_t$ is connected to the output, $Y_t$. Each time step, $x_{t-\tau}$ of the input $X_t$ is given to a single LSTM cell. The input is concatenated to the hidden memory output of the preceding LSTM cell, $h_{t-1}$, forming $x'_t$. The input $x_t'$ and the internal weights of the current LSTM cell change the information $c_{t-1}$ from the previous cell by propagating, modifying or blocking the information. Finally, a new hidden state $h_t$ is created by first transforming $x_t'$ with one of the LSTMs internal neural networks and then combining it with the new memory state $c_t$. The final output of the cell, the hidden state $h_t$, is obtained by the internal combination of the cell input $x_t'$ and the memory of previous cells, $c_t$, using an internal neural network.

The final hidden state vector, $h_6$, is then fed to a classical, fully connected feed-forward neural network, where the input is transformed to a vector of the size of the target vector $Y_t$. Throughout the rest of the paper, we will refer to this model as the `LSTM-NN model'.

The model is trained using the RMSProp method, an unpublished adaptive learning rate method proposed by Goeff Hinton (see \url{http://www.cs.toronto.edu/~tijmen/csc321/slides/lecture_slides_lec6.pdf}). As with any other neural network based technique, LSTMs ``learn" by iteratively adjusting the weights of the inter-neuron connections. The iterative process is similar to the Newton-Raphson method, adjusting the free parameter of the model by gradual changes based on the derivative of the error between the real output and the forecast output \citep{hecht1992theory, lecun2012efficient}. LSTMs are composed of a chain of neural networks that learn together. This requires a special optimization method called RMSprop. RMSprop ensures that the error is correctly propagated backwards through all the chain of neural networks that compose the LTSM. The method has been used successfully for training LSTMs in generating complex sequences \citep{graves2013generating}.

The PyTorch library provides an implementation of this method. The error criterion of the model was the mean squared error loss:
\begin{equation}
	MSE = \frac{1}{N} \sum_{i=1}^N (M_i - O_i)^2.
\end{equation}
When the model is trained with a training set, the error on this set will be smaller for every iteration (or epoch). In order to prevent over-fitting, the process where the model starts memorizing the training set instead of learning the training set, after every epoch, the performance of the model on the validation set is checked. When the model performance stops improving on the validation set is a good indicator that the model is starting to over-fit on the training set, and we can stop the training of the model. This is the classic early stopping method.

\subsubsection{Parameters of the model}

When training a neural network, there are many parameters that can be chosen that have an impact on the performance. However, finding the optimal value for these parameters is problem-dependent, the so-called No Free Lunch theorem \citet{wolpert1997no}, and finding the optimal values is a computationally exhaustive task. In our model, there are three sets of hyperparameters. The first set is intrinsic to the architecture of our model itself, the second set to the learning method, and the third set to the training method. The hyperparameters were obtained by manually tweaking their values over the course of 15 to 20 runs and evaluating their performance using 7-fold CV. We found that for our particuliar case, we observed no significant changes in the accuracy of the model caused by the tweaking. Because the model seemed robust under the tweaking, we decided to not do an exhaustive search for the optimal values of the hyperparameters. We now provide an overview of the important hyperparameters and the values we gave them. A final list of the parameters are summed up at the end of this section.

In our model itself, the LSTM has a few hyperparameters that have impact on its performance. The first is the number of neurons in the hidden layer of the LSTM. This number must be large enough to ensure it can encode the process behind the data, but not too large to prevent over-fitting and plain computational cost. This number has been determined by performing CV, and we found the best performance to be around 50 neurons. Next there is the number of layers in the LSTM. Multiple LSTM's can be stacked on top of each other, where the first LSTM gives the intermediate hidden states (see Figure \ref{fig:model:model}) as input for the second LSTM, and so on. This increases the complexity and computational cost of the model. In our search we tested using multiple layers of LSTMs, but did not find any significant performance increase and set the number of layers to 1. Finally, there is the option to make the LSTM bidirectional, where the model has access to both future and past states. But this is unnecessary in the context of the Dst forecast, thus this has been set to false.

The RMSProp method has a lot of tweakable parameters, but we will focus on the two most important parameters, the learning rate and the momentum. The learning rate is the most important parameter, and controls how strongly the model weights will be changed by the error gradient. A too large learning rate might cause unstable learning, with the performance swinging widely during training and preventing convergence. A too small learning rate might cause the model to barely change and converge too slowly or not at all. The momentum parameter will affect the learning rate parameter throughout the training process, and will accelerate the training process. Most often, the momentum parameter is chosen close to 1. We found that setting the learning rate to 0.0003 and the momentum to 0.8 gave us the best performance, with stable convergence of the training error. 

Finally, when training the model, we can also set a few parameters that can affect the performance. There are two parameters that are important: the number of epochs and the batch-size of the training set. The number of epochs decide how many times we loop over the full training set for training. This number must be large enough that the model has time to converge to the ideal solution before the training is stopped by the classic early stopping method. The batch-size determines how many samples of the training data are given to the model before the error is computed and backpropagation is applied. Setting the batch-size to one would corresponds with so-called online learning, where the model is trained separately on every sample. The opposite is offline learning, i.e. setting the batch-size to the size of the training set, so the model is optimized on the accumulated error over the complete training set. Offline learning is almost never used as it fails to learn more outlying cases, and online learning is more prone to over-fitting. Using a small batch-size is typically recommended. We found that setting the batch-size to 64 gave a fast convergence and did not have a large impact on the performance.

Finally, we sum up the parameters of the model:
\begin{itemize}
	\item[1.] LSTM hyperparameters
	\begin{itemize}
		\item Number of hidden layer neurons: 50 
		\item Number of layers: 1
		\item Bidirectional: False
	\end{itemize}
	\item[2.] RMSProp hyperparameters:
	\begin{itemize}
		\item Learning rate: 0.0003
		\item Momentum: 0.8
	\end{itemize}
	\item[3.] Training hyperparameters
	\begin{itemize}
		\item Number of epochs: 30
		\item Training set batch-size: 64 
	\end{itemize}
\end{itemize}

\subsection{Results and discussion}
\label{sect:exp:Results}

The LSTM-NN model is now evaluated using the defined metrics and baseline model of Subsection \ref{sect:exp:eval}. The evaluation is discussed and a comparison of the model to some of the latest forecasting models is made. Finally, the forecast is visually observed.

The first analysis examines whether the LSTM-NN model performs better than the persistence model defined in Section \ref{sect:exp:eval:baseline}. Table \ref{table:LSTMvsPers} displays the results from the metrics defined in Subsection \ref{sect:exp:eval:metrics}, applied on the forecast of both the LSTM-NN model and the persistence model. The LSTM-NN model was found to overall perform better than the persistence model. Only the linear relation of the persistence model is consistently better for every forecasting horizon compared to the LSTM-NN model. 

Taking a closer look at the remaining metrics, it seems that the MAE of the LSTM-NN model and the persistence model are similar. It increases with the forecasting horizon, and always remains smaller than the RMSE. The correlation, linear model parameters, and prediction efficiency of the LSTM-NN model are close to that of the persistence model for forecasting horizon $t+1h$ and $t+2h$; however, its accuracy quickly disappears for later forecasting horizons. This would indicate that the persistence model could serve as a strong benchmark for nowcasts of the Dst index. Taking into account the results reported in Table \ref{table:LSTMvsPers}, we conclude that using the more complicated LSTM-NN model will result in better forecasts. In particular, forecasts made at forecasting horizon $t+3h$ to $t+6h$ show significant improvement in accuracy compared to the persistence model.
 
\begin{table}[htbp!]
	\centering
	\caption{Evaluation of the LSTM-NN model and the persistence model with the metrics from Section \ref{sect:exp:eval:metrics}. The values in bold show where the persistence model performed better than the LSTM-NN model. \label{table:LSTMvsPers}}
	\begin{tabular}{@{}lccccccc@{}} 
		\toprule
		Forecasting horizon & RMSE & R & A & B & MAE & ME & PE \\
		& (nT) & & & (nT) & (nT) & (nT) &  \\ \hline 
		\textit{LSTM-NN model} \\ 
		\quad $t+1h$ & 3.731 & 0.980 & -0.166 & 0.960 & 2.391 & 0.318 & 0.960 \\
		\quad $t+2h$  & 5.689 & 0.953 & -0.770 & 0.915 & 3.820 & 0.271 & 0.907 \\
		\quad $t+3h$  & 7.155 & 0.924 & -1.438 & 0.866 & 4.823 & 0.198 & 0.853 \\
		\quad $t+4h$  & 8.172 & 0.899 & -1.963 & 0.826 & 5.479 & 0.161 & 0.808 \\
		\quad $t+5h$  & 8.926 & 0.878 & -2.358 & 0.793 & 5.929 & 0.168 & 0.771 \\
		\quad $t+6h$  & 9.566 & 0.859 & -2.769 & 0.761 & 6.280 & 0.143 & 0.737 \\
		\textit{Persistence model} \\
		\quad $t+1h$ & 4.745  & 0.974 & \ 1.884  & \textbf{0.975} & 3.265 & 2.179 & 0.935 \\
		\quad $t+2h$  & 6.853  & 0.934 & \ 0.101  & \textbf{0.935} & 4.400 & 0.875 & 0.865 \\
		\quad $t+3h$  & 8.998  & 0.895 & \ 1.571  & \textbf{0.898} & 5.884 & 2.798 & 0.767 \\
		\quad $t+4h$  & 9.913  & 0.860 & -0.652 & \textbf{0.864} & 6.300 & 0.979 & 0.717 \\
		\quad $t+5h$  & 10.937 & 0.829 & -1.487 & \textbf{0.834} & 6.943 & 0.507 & 0.655 \\
		\quad $t+6h$  & 11.864 & 0.799 & -1.592 & \textbf{0.805} & 7.464 & 0.742 & 0.594 \\
		\bottomrule
	\end{tabular}
\end{table}

Next we compare the LSTM-NN model to the models reported in the work of \citet{Gruet2018} and \citet{Lazzus2017}. Both publications present a neural network trained on OMNIWeb data used to forecast the Dst index. The model by \citet{Gruet2018} also makes use of LSTM modules in their model, while the model of \citet{Lazzus2017} consists of a feed-forward neural network instead of a RNN, trained with a particle-swarm optimization method. The performance of their models were evaluated with the RMSE and Pearson correlation coefficient, and are summarized in Table \ref{table:compareResults}, together with the results from the LSTM-NN model and the persistence model. 

\begin{table}[!htbp]
	\centering
	\caption{The RMSE and Pearson linear correlation coefficient of the persistence model and LSTM-NN model compared to the models of \citet{Gruet2018} and \citet{Lazzus2017}. The confidence intervals were obtained from the cross-validation experiment, detailed in Section \ref{sect:exp:Results} and shown in Figure \ref{fig:cv_var}.} \label{table:compareResults}
	\begin{tabular}{@{}lcccc@{}} 
		\toprule Forecasting horizon & Persistence & LSTM-NN & Gruet et al. & Lazz\'us et al. \\ & & & (2018) & (2017) \\ \hline
		\textit{Correlation} \\
		\quad $t+1h$ & $0.974 \pm 0.003$ & $0.980 \pm 0.008$ & 0.966 & 0.982 \\
		\quad $t+2h$ & $0.934 \pm 0.009$ & $0.953 \pm 0.010$ & 0.946 & 0.949 \\
		\quad $t+3h$ & $0.895 \pm 0.015$ & $0.924 \pm 0.013$ & 0.923 & 0.918 \\
		\quad $t+4h$ & $0.860 \pm 0.019$ & $0.899 \pm 0.017$ & 0.902 & 0.887 \\
		\quad $t+5h$ & $0.829 \pm 0.023$ & $0.878 \pm 0.019$ & 0.882 & 0.858 \\
		\quad $t+6h$ & $0.799 \pm 0.026$ & $0.859 \pm 0.021$ & 0.865 & 0.826 \\
		\textit{RMSE (nT)} \\
		\quad $t+1h$ & $4.75 \pm 0.47$ & $3.73 \pm 0.78$ & 5.34 & 4.24 \\
		\quad $t+2h$ & $6.85 \pm 0.85$ & $5.69 \pm 0.59$ & 6.65 & 7.05 \\
		\quad $t+3h$ & $9.00 \pm 1.10$ & $7.16 \pm 0.55$ & 7.86 & 8.87 \\
		\quad $t+4h$ & $9.91 \pm 1.39$ & $8.17 \pm 0.59$ & 8.86 & 10.44 \\
		\quad $t+5h$ & $10.94 \pm 1.56$ & $8.92 \pm 0.70$ & 9.59 & 11.65 \\
		\quad $t+6h$ & $11.86 \pm 1.74$ & $9.57 \pm 0.83$ & 10.24 & 13.09 \\
		\bottomrule
	\end{tabular}
\end{table}

However, before we can do quantitative comparison of the LSTM-NN model with the two other models presented in the literature, we have to keep in mind the following problem: direct comparison of two different neural network models is considered bad practice when the sets used to train and test both models are not identical, as also stated by \citet{Lazzus2017}.Because we are unable to recreate the exact same training and test set used by Gruet and Lazz\'us, we instead will quantify the impact that the choice of training and test set has on the performance of the LSTM-NN model. By performing 10-fold CV, we can measure the variance in performance caused by the choice of training and test set, as 10-fold CV will replicate the effect of training the model for 10 different choices of training and test set.
	
\begin{figure}
	\centering
	\includegraphics[width=\textwidth]{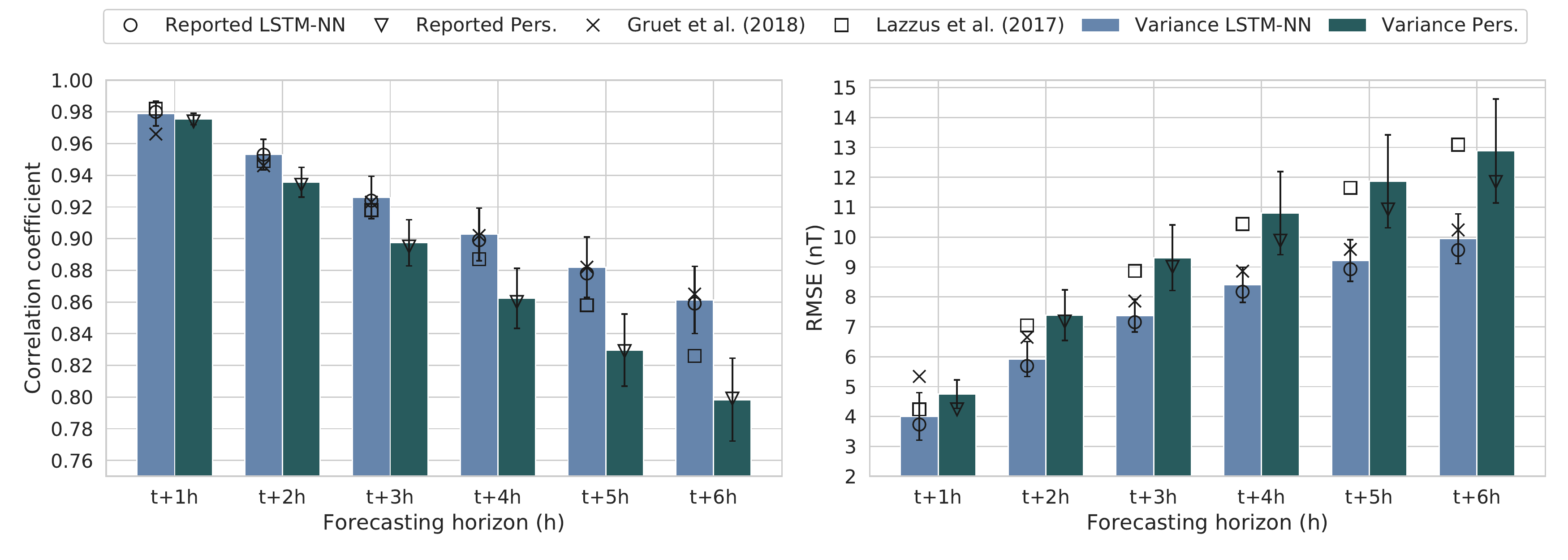}
	\caption{Using 10-fold cross validation, the LSTM-NN model and persistence model were evaluated for 10 different training and test sets with the RMSE and correlation coefficient to determine the impact of the choice of set on the performance. The left figure displays the correlation coefficient, whose values we want to have as close to 1 as possible. The right figure displays the RMSE, whose values we want to minimize. The circle and triangle display the values reported in Table \ref{table:LSTMvsPers}, the bars and error bars respectively display the mean and variance of the performance of the 10-fold cross-validation, and the square and cross display the performance of the models reported by \citet{Lazzus2017} and \citet{Gruet2018}, respectively. All of these values are also summarized in Table \ref{table:compareResults}.}
	\label{fig:cv_var}
\end{figure}

From this we can learn two things. The first has already been discussed in Section \ref{sect:exp:preprocessing}, namely a way to quantify the effect of our final choice of training, validation and test set. By choosing an ideal training and test set, it is possible to cause an artificial improvement of the model performance. Computing the variance caused by the choice of training and test makes it possible to determine if the reported performance is an outlier, or expected. The results are shown in Figure \ref{fig:cv_var}. The average performance of the LSTM-NN and the persistence model from the CV is indicated by the blue and dark green bars, and the standard deviation on the performance is indicated by the error bars. The uncertainty interval obtained from this experiment has also been included in Table \ref{table:compareResults}. We find that for both RMSE and correlation, the reported performance of the LSTM-NN and the persistence model lies inside the variation, indicating that there is no artificial improvement of the results by choosing an ideal training and test set.

The second reason to do this experiment was to be able to perform a qualitative comparison of the LSTM-NN with the model of Gruet and Lazz\'us. Their reported values have also been added to Figure \ref{fig:cv_var}. Let us first look at the results for the correlation in the left bar chart. The model of \citet{Lazzus2017} has a performance that is comparable to that of the LSTM-NN model for forecasting horizons $t+1h$ to $t+4h$. However, for later forecasting horizons, the LSTM-NN performs significantly better. The model by \citet{Gruet2018} is outperformed by the LSTM-NN for the forecasting horizon $t+1h$, but performance equivalently for all the later forecasting horizons.

Looking at the RMSE, we see that the model by Lazz\'us has comparable performance for forecasting horizon $t+1h$. However, for later forecasting horizons the LSTM-NN model is significantly better. This seems to agree with the observation made by \citet{Gruet2018}, stating that the choice of an LSTM module in the model architecture improves the accuracy of the forecast. Looking at the performance of Gruet, the LSTM-NN shows a significant improvement for forecasting horizons $t+1h$ to $t+2h$, but is equivalent in performance for later times.

In conclusion, our model seems to be comparable to that of Lazz'us for forecasting horizons $t+1h$ and $t+2h$, but outperforms it for later times. Our model shows some improvement over that of Gruet for forecasting horizons $t+1h$ and $t+2h$, but is otherwise comparable to theirs. 

\begin{figure}
	\centering
	\includegraphics[width=\linewidth]{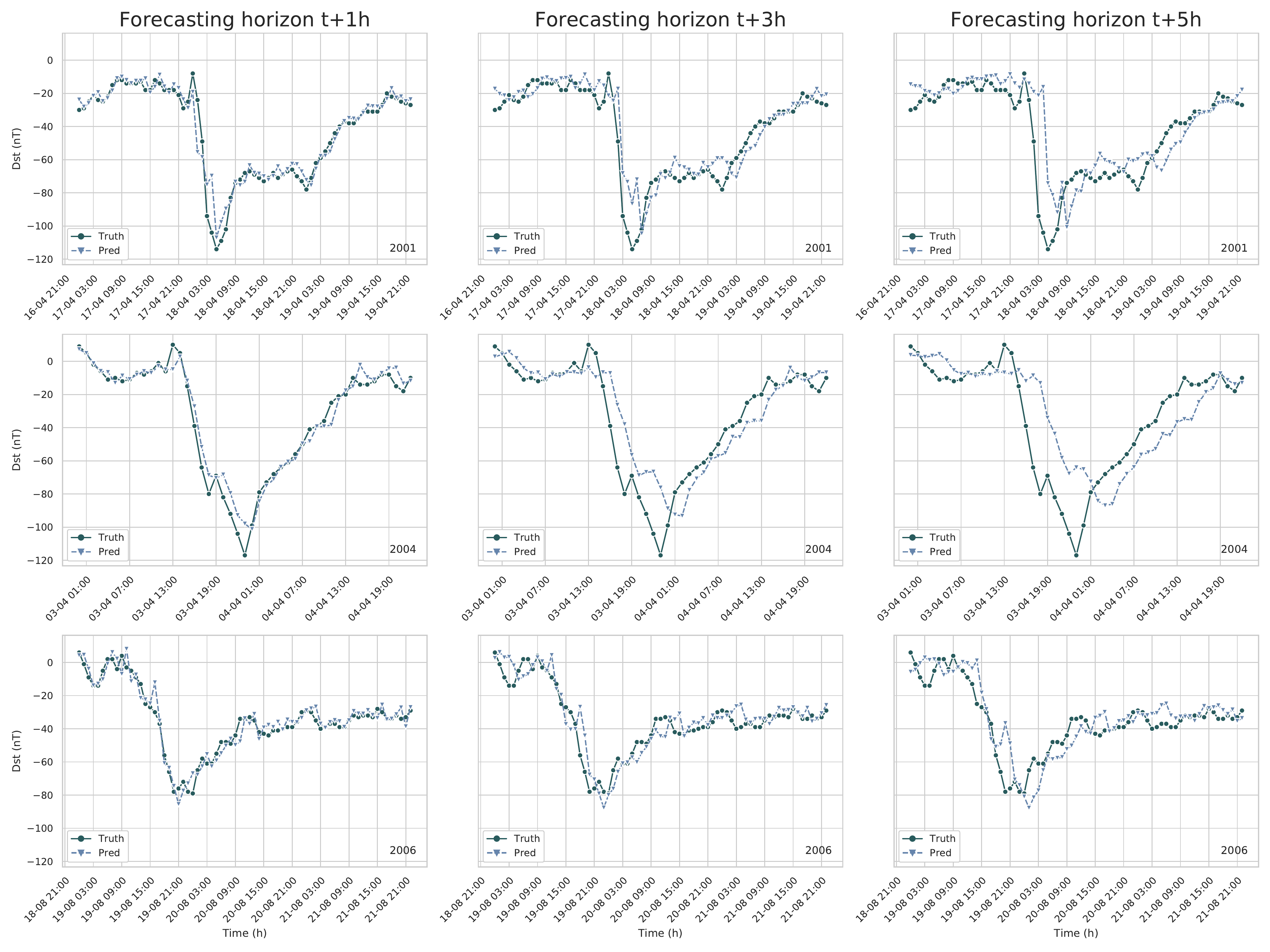}
	\caption{Forecast of the LSTM-NN model for three separate geomagnetic storm events. Each plot shows the observed Dst index (Truth) and the forecast Dst index (Pred). Every row shows the same storm, and every column corresponds to a different forecaseting horizon. Notice that the LSTM-NN model systematically forecasts the main phase of the storm too late by a number of hours equivalent to the forecasting horizon.}
	\label{figs:results_model}
\end{figure}

Finally, a visual observation of the forecast is analyzed. Figure \ref{figs:results_model} displays three geomagnetic storms contained in the test set, together with the forecast of the LSTM-NN model for forecasting horizon $t+1h$, $t+3h$ and $t+5h$. The first column displays the $t+1h$ forecast, and seems to be an almost perfect prediction of the storm. However, the forecast of the Dst-index for forecasting horizon $t+3h$ and $t+5h$, displayed in column 2 and 3 of Figure \ref{figs:results_model}, shows a distinct delay in the forecasting of the main phase. Take for example the prediction at forecasting horizon $t+5h$: the sudden offset of the storm is predicted 5 hours too late. 

This brings us to the main problem of this paper. The purpose of the experiment was to create a LSTM-NN model that forecasts the Dst-index with the same accuracy and correlation as other presented architectures. We managed to create such a model, but, when visually inspecting the forecast, it was observed that there is a distinct time shift between forecast and observation. If geomagnetic storms are forecast only when they start, it means the LSTM-NN model will not give us any more information than the persistence model. While it is not possible to say that the models from \citet{Gruet2018} and \citet{Lazzus2017} also have this problem, we believe that one should pay close attention to this problem and ensure it does not happen.

An additional problem that most modern machine learning techniques have to face is that rare events can not be properly forecasted. Neural networks learning by gradient descent requires that patterns show up frequently in the data. In order to forecast dangerous super-storms, like the 2003 Halloween or the Carrington events, the networks must have to learn to identify them. From 2001 to 2016 there are only 100 entries of the Dst index recorded with values bellow -200nT (including consecutive hours of individual storms). Possible solutions to this issue can be of four types: 1) data augmentation by duplication, where months with high number of storms are used multiple times in a singular epoch, b) generative data augmentation, where a second machine learning technique, like auto-encoders or generative adversarial networks, is used to generate artificial storms, 3) augmentation by computer simulations, using 3D models of the interaction of the solar wind and the magnetosphere of the planet to artificially generate data with large storms, and 4) multi-tier machine learning architectures, where multiple models specialize in the detection of different types of inputs and storms strengths. These solutions are out of the scope of the present paper but will be studied in a future work.

\section{The Dynamic Time Warping Method}
\label{sect:DTW}

Section \ref{sect:Experiment} revealed that the LSTM-NN model failed to give an accurate forecast of the Dst index, and in particular geomagnetic storms, despite the evaluation of the model indicating that the model should have a high accuracy and correlation. This problem has also been observed by \citet{wintoft2018evaluation} and \citet{Stepanova2000}, but not by other similar forecasting models. It is often unclear whether or not this was overlooked or if the forecasting model did not have this problem. Wintoft and Wik were able to detect this time shift by manually shifting their forecast in time and analysing the correlation coefficient between shifted forecast and observation, while Stepanova and P\'erez visually observed this time shift.

Because this phenomena seems detrimental in the evaluation of a forecasting model, we propose a new method, which we will name the `warping measure'. This measure is more capable of quantifying the time shift between model and observation, and is based on the Dynamic Time Warping (DTW) algorithm \citep[see][]{berndt1994using}, a method that measures the relative similarity between two time series. At the very least, we expect the warping measure to be able to detect the forecast made by a persistence model. What follows first is a brief overview of the DTW algorithm, followed by the modifications we made to tailor the algorithm to our specific problem.
\hphantom{f}

\subsection{Dynamic Time Warping}
\begin{figure}[ht]
	\centering
	\includegraphics[width=\textwidth]{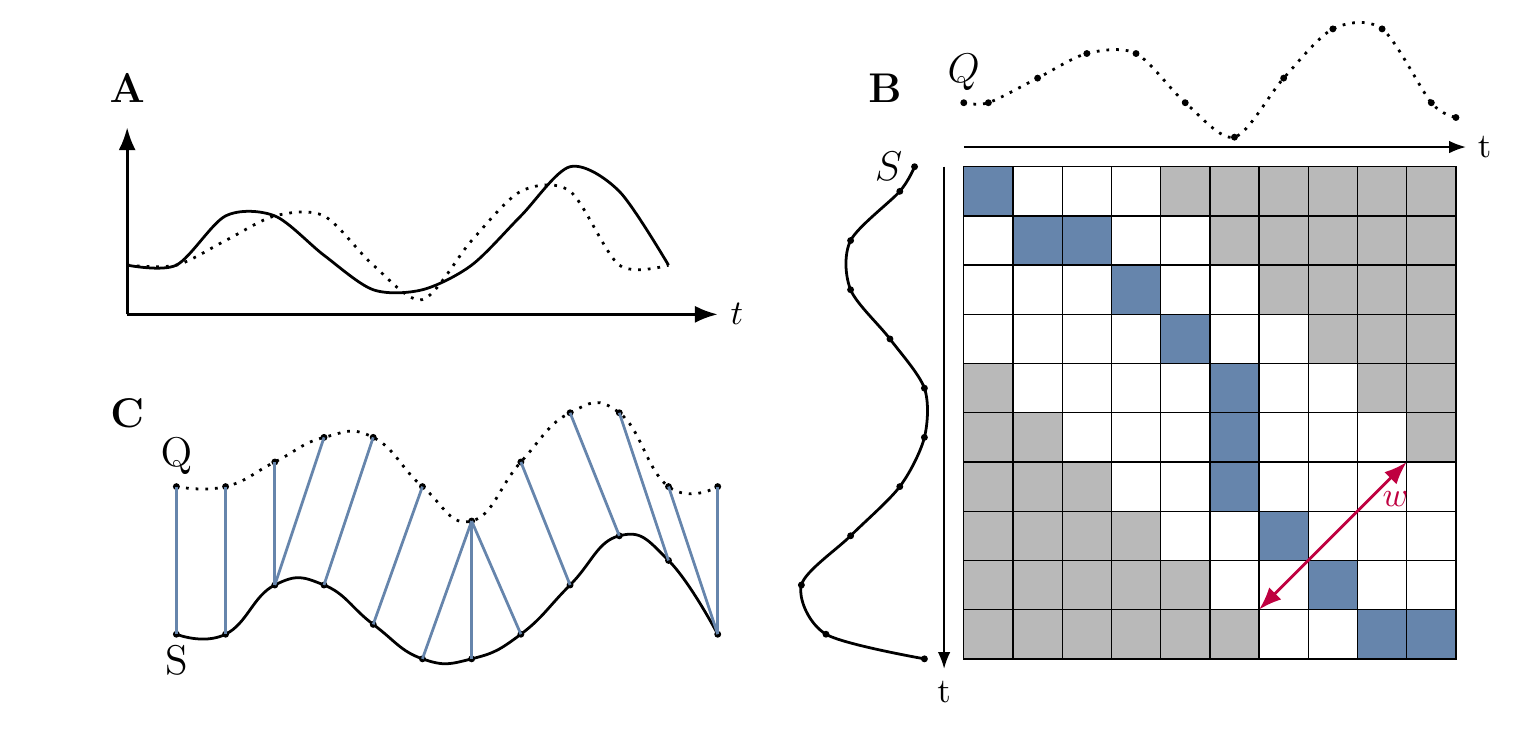}
	\caption{\textbf{(A)} shows two time series we want to compare. \textbf{(B)} illustrates the cumulative distance matrix, together with a warping window $w$ and the ideal warping path $P$ in blue. \textbf{(C)} illustrates the warping path $P$ aligning the two time series.} \label{fig:dtw_algorithm}
\end{figure}

The DTW algorithm is a method first developed for speech recognition and is now commonly used in the fields of economics, biology, and database analysis \citep[see e.g.][]{skutkova2013classification, wang2012similarity}. DTW is mainly used as a measure to investigate how much a sample time series matches or is contained in a target time series. The strength of DTW is that it can compare two time series even though they might be shifted or stretched in time, which is a property that is essential to our goal. This section summarizes the algorithm developed by \citet{berndt1994using}. A visualization of this algorithm is shown in Figure \ref{fig:dtw_algorithm}. Take two time series, Q and S, of length $n$ and $m$, respectively.
\begin{equation}
	Q = [q_1, q_2, \dots, q_i, \dots, q_n],
\end{equation}%
\begin{equation}
	S = [s_1, s_2, \dots, s_j, \dots, s_m]. 
\end{equation}
The DTW algorithm first constructs a distance between these two time-series by placing them in an $n \times m$ grid. Each grid point $(i,j)$ then corresponds to an alignment of $q_i$ and $s_j$. An alignment is given a cost by a distance function $d(q_i, s_j)$. The distance function can be chosen freely, and for our case the Euclidean distance function, $d(x, y) =\sqrt{(x - y)^2}$ is used. 
The DTW algorithm then searches for a path (the so-called warping path) $P$ in this grid that minimizes the sum of said distance. The warping path $P$ can be defined as:
\begin{equation}
	P = [p_1, p_2, \dots, p_K], \text{ with } \max(m,n) \le K < m + n - 1,
\end{equation}
where each point $p_k$ corresponds to a grid point $(i,j)_k$. The path must then minimize the cost function, so 
\begin{equation}
	DTW(Q, S) = \min \sqrt{\sum^K_{i=1} p_k },
\end{equation}
and must hold to the following conditions:
\begin{enumerate}
	\item Boundary conditions: the beginning and the end of the sequences are matched;
	\item Continuity: there are no gaps, every point is mapped to at least one other point;
	\item Monotonicity: the points are ordered in time, $i_{k-1} \le i_k$ and $j_{k-1} \le j_k$;
	\item Warping window $w$: an optional constraint that sets the maximum distance in time between $i_k$ and $j_k$ to w: $|i_k-j_k| \le w$.
\end{enumerate}

\begin{figure}[ht]
	\centering
	\includegraphics[width=\textwidth]{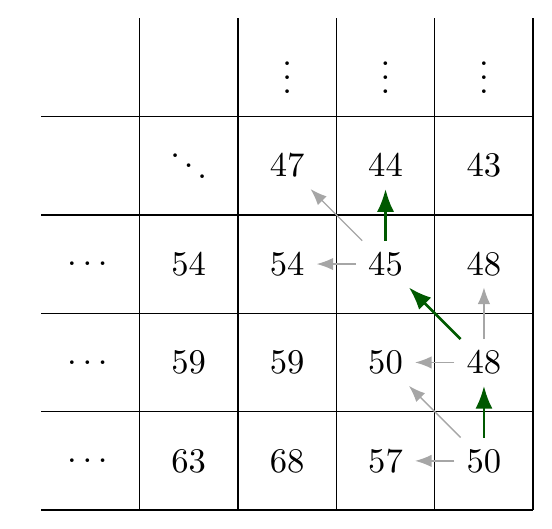}
	\caption{Illustration of how the warping path is determined from the distance matrix $D$.} \label{fig:dtw_path}
\end{figure}

In order to find this optimal path, the following dynamic programming technique can be used. Starting at point $(1,1)$, the cumulative distance $\Delta$ at each grid point is computed by the following recursive equation:
\begin{equation}
	\Delta[i, j] = d(q_i, s_j) + \min(\Delta[i-1, j-1], \Delta[i-1, j], \Delta[i, j-1]).
\end{equation}
Once all the cumulative distances are computed, the optimal warping path can be found by starting at the point $(n,m)$ and tracing backwards in the grid, taking the smallest value each time. This is displayed in Figure \ref{fig:dtw_path}.

A warping window constraint can be added on the algorithm. This window will change the warping cost and warping path $P$. Let $w \in \mathbb{N}$ be the warping window, then 
\begin{equation}
	\forall p_k \in P, \quad |i_k - j_k| \le w.    
	\label{eq:dtw:warpingconstraint}
\end{equation}
Faster and better implementations of this algorithm exist \citep[see e.g.][]{keogh1999scaling, keogh2005exact, lemire2009faster, salvador2007toward}, but they are outside the scope of this text. 

\subsection{The warping measure}
It should be mentioned that the DTW algorithm does not satisfy the necessary properties to be a metric. For example, it is easy to see that the algorithm does not satisfy the triangle inequality. Consequently, this method will be called a measure, and not a metric. 

\begin{figure}
	\centering
	\includegraphics[width=\textwidth]{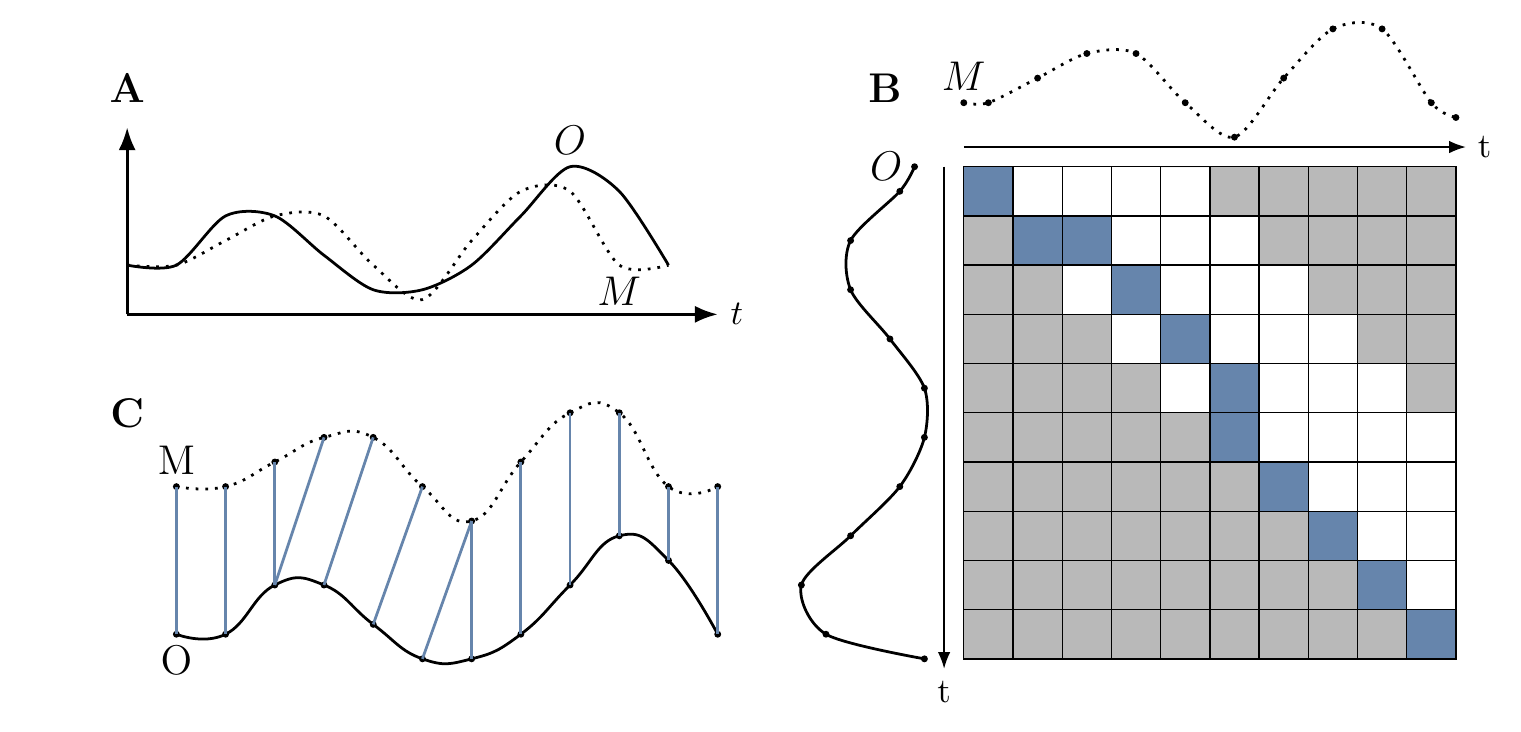}
	\caption{Overview of the warping measure. The two time series M and O are compared for alignment. However, the warping algorithm is adapted with a new window $w$ that prevents comparing values of $M$ with values of $O$ that lie in the future, as seen in b). Figure c) shows the alignment from b) from which the measure is determined. Notice that there is no alignment from $M$ to points in $O$ that lie in the relative future of $M$.}
	\label{fig:dtw_measure}
\end{figure}

This measure does not make use of the warping cost, and instead uses the information contained in the warping path $W$. The measure is then able to determine how exactly a forecast time series is shifted in time in comparison to the true or observed time series. Take the two time series $M$ and $O$, where $M$ is the predicted time series and $O$ the true time series.
\begin{equation} 
	M = [m_1, m_2, \dots, m_i, \dots, m_n],
\end{equation}
\begin{equation} 
	O = [o_1, o_2, \dots, o_j, \dots, o_m].
\end{equation}
The DTW algorithm is applied on these time series, giving a cost matrix $D$ of dimensions $n \times m$. The warping constraint defined in Equation \eqref{eq:dtw:warpingconstraint} is applied, and $w$ is set equal to the forecasting horizon time.
However, an additional constraint is included: the warping window is restrained such that the algorithm only compares the prediction $M$ at $t+p$ with the observations $O$ from time $t$ to $t+p$, i.e. predictions are not compared to observations that are in the future. Applying this constraint can be done as a modification of the warping constraint defined in Equation \eqref{eq:dtw:warpingconstraint}:
\begin{equation}
	\text{Let } m_i \text{ be the modeled value, } o_j \text{ the observation, then } w \ge i - j \ge 0.
\end{equation} 
This is also illustrated in Figure \ref{fig:dtw_measure}. After computing the warping path, we take each step $p_k = (i_k, j_k)$ and compute what we define as the warp value:
\begin{equation}
	\Delta t = |i_k - j_k|, \text{ with } \Delta t \in [0, 1, \dots, w].
\end{equation}
Finally, a histogram is taken from all the different values of $\Delta t$. The percentages reflect how time series $M$ is shifted compared to time series $O$. We now present the results of this measure applied to the persistence model prediction and the LSTM-NN model prediction. 

\hphantom{ }

\subsection{Results}

\subsubsection{DTW measure applied to the persistence model}

The warping measure is first applied to the forecast of the persistence model. The persistence model can be seen as the textbook example for this algorithm. Assuming that the persistence model is set as follows:
\begin{equation}
	Dst(t+p) = Dst(t), \quad p \in \mathbb{N},
\end{equation}
then the algorithm should detect that almost all of the forecast values are shifted with a time $p$ compared to the actual observation. The algorithm will not detect 100\% of the values to be shifted with $p$, because of the constraint in the DTW algorithm that forces the beginning and the end points of the two time series to match, as discussed in Section \ref{sect:DTW}. 

\begin{table}[htbp]
	\centering
	\caption{The row-normalized fractions of the warping measure on the persistence model. The algorithm detects that the persistence model is shifted with the expected number of hours.} \label{table:persistence_dtw_result}
	\begin{tabular}{@{}llllllll@{}} 
		\toprule
		Forecast horizon & \multicolumn{7}{c}{Time shift} \\
		\cmidrule{2-8}
		& 0h    & 1h    & 2h    & 3h    & 4h     & 5h    & 6h \\
		\hline
		$t+1h$ & 0.003 & \textbf{0.997} & 0.0   & 0.0   & 0.0   & 0.0   & 0.0   \\
		$t+2h$ & 0.003 & 0.003 & \textbf{0.994} & 0.0   & 0.0   & 0.0   & 0.0   \\
		$t+3h$ & 0.004 & 0.003 & 0.003 & \textbf{0.991} & 0.0   & 0.0   & 0.0   \\
		$t+4h$ & 0.003 & 0.003 & 0.003 & 0.003 & \textbf{0.988} & 0.0   & 0.0   \\
		$t+5h$ & 0.004 & 0.003 & 0.003 & 0.003 & 0.003 & \textbf{0.984} & 0.0   \\
		$t+6h$ & 0.004 & 0.003 & 0.003 & 0.003 & 0.003 & 0.003 & \textbf{0.981} \\
		\bottomrule
	\end{tabular}
\end{table}

The persistence model is applied to the test set defined in Section \ref{sect:exp:data}, and the resulting warp values are shown in Table \ref{table:persistence_dtw_result}. The results confirm our expectations, where except for a few percentile, all the values are detected to be shifted by the forecasting horizon. 

\begin{figure}[ht]
	\centering
	\includegraphics[width=\textwidth]{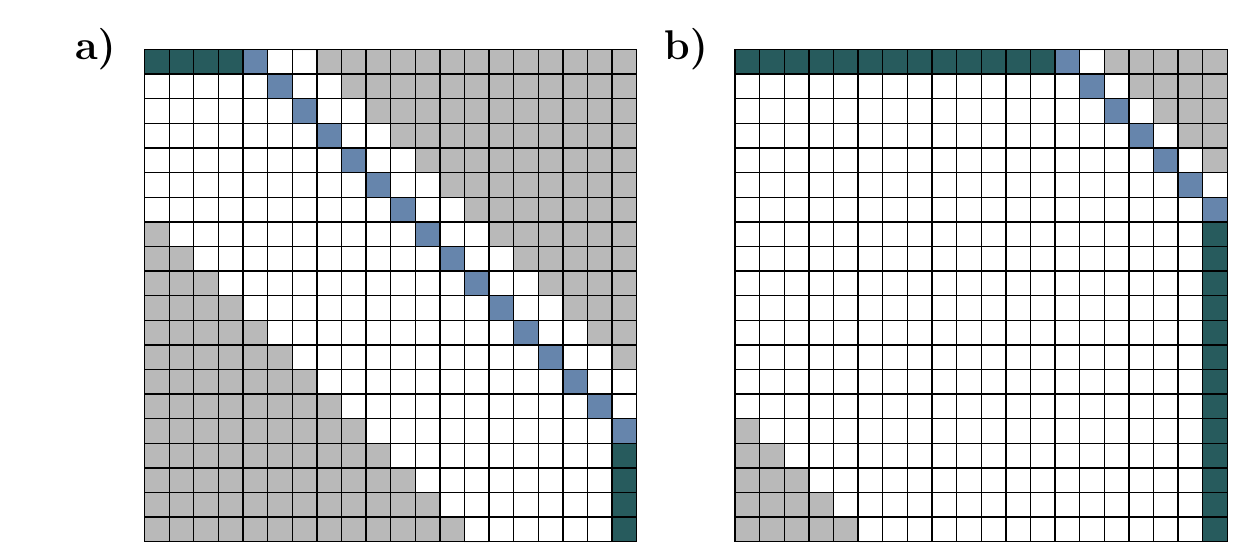}
	\caption{An illustration of a potential problem with the measure. Panel \textbf{(A)} illustrates the result for a persistence model with a small time-shift. Panel \textbf{(B)} illustrates the case of a persistence model with a large time shift. When the time series is too small, the counts will be dominated by the green-colored block, while the actual truth will appear very small due to normalization.} \label{fig:dtw_problem}
\end{figure}

One potential problem that can arise is when the time shift in the two compared time series is very large. First, the algorithm will take longer to run as the window-size $w$ needs to be much larger. Second, because the boundary conditions require the beginning and end of both sequences to match, if the time series is too short, the algorithm might give a shifted results. Take the extreme example shown in Figure \ref{fig:dtw_problem}, showing both a persistence model with respective time shift $s_1$ and $s_2$. If the time shift is very large and the time series is small, the number of values counted to have shift $s_2$ are made insignificant due to the path also including the boundary condition. This effect is already starting to show in the final row of Table \ref{table:persistence_dtw_result}, where around 2\% of the shifted values are artificial due to the boundary constraint. Potential changes to the algorithm that could account for this problem is a topic for future work.

\subsubsection{DTW applied to the LSTM-NN model}

The normalized values of the DTW measure applied to the LSTM-NN model presented in Subsection \ref{sect:exp:model} can be seen in Table \ref{table:dtw_lstm}. The highest percentages are located on the offset diagonal, identical to the results of the persistence model. As discussed before, this indicates that a shift in time exists between the observations and the model predictions. This confirms that our observation of the results discussed in Section \ref{sect:exp:Results} are happening throughout the whole time series. We notice that the second highest percentage of each row is located on the diagonal, indicating that the model is actually capable of providing some accurate prediction of the Dst for one hour into the future, similar to the observations of \citet{wintoft2018evaluation} and \citet{Stepanova2000}.

\begin{table}[!htbp]
	\centering
	\caption{The row-normalized fractions of the DTW measure on the LSTM-NN model. The results are in agreement with the visual inspection of the model: for each forecasting horizon time, the highest percentage is located at the corresponding time shift.}  \label{table:dtw_lstm}
	\begin{tabular}{@{}llllllll@{}}
		\toprule
		Forecast horizon & \multicolumn{7}{c}{Time shift} \\ \cmidrule{2-8}
		& 0h    & 1h    & 2h    & 3h    & 4h     & 5h    & 6h \\ \hline
		$t+1h$ & 0.352 & \textbf{0.578} & 0.045 & 0.015 & 0.006 & 0.003 & 0.002 \\
		$t+2h$ & 0.115 & 0.334 & \textbf{0.428} & 0.069 & 0.03  & 0.015 & 0.009 \\
		$t+3h$ & 0.074 & 0.113 & 0.287 & \textbf{0.355} & 0.09  & 0.047 & 0.034 \\
		$t+4h$ & 0.068 & 0.066 & 0.115 & 0.249 & \textbf{0.309} & 0.11  & 0.083 \\
		$t+5h$ & 0.073 & 0.054 & 0.073 & 0.117 & 0.225 & \textbf{0.28}  & 0.178 \\
		$t+6h$ & 0.079 & 0.052 & 0.06  & 0.079 & 0.117 & 0.215 & \textbf{0.397} \\
		\bottomrule
	\end{tabular}
\end{table}

\section{Discussion}
\label{sect:Discussion}

\subsection{Dst index analysis}

\begin{figure}[htbp]
	\centering
	\includegraphics[width=\textwidth]{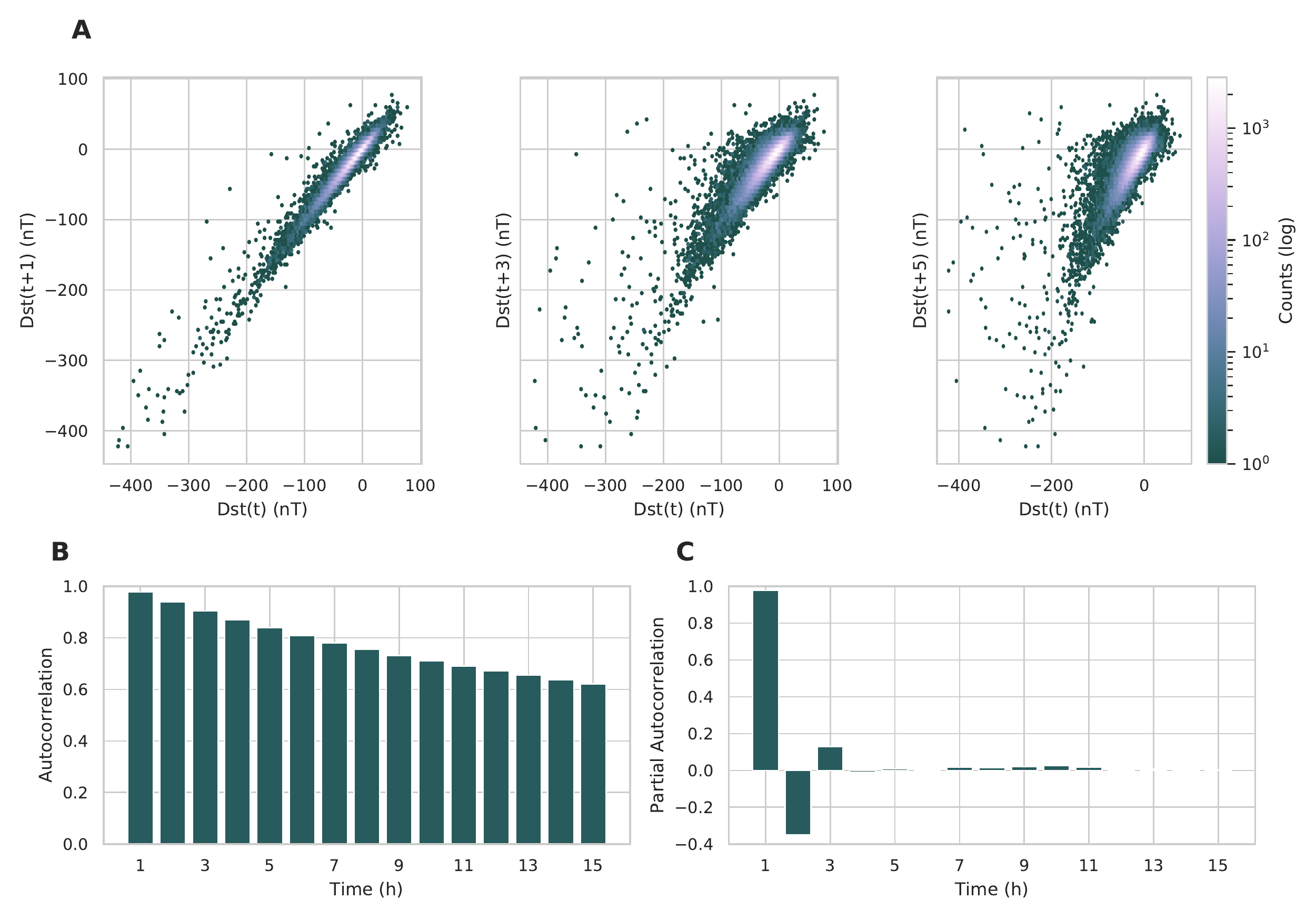}
	\caption{Panel \textbf{(A)}displays three lag plots of the Dst, with lag times of 1, 3 and 5 hours. The color of each point represents the how many times this value was encountered in the data. Notice the strong linear behavior for all three lag times. Panel \textbf{(B)} shows the autocorrelation of the Dst index for up to 15 hours of lag. In panel \textbf{(C)}, the partial autocorrelation of the Dst index. Notice that there is almost no more correlation past 2h.} \label{fig:dst_correlation}
\end{figure}

What follows is a statistical analysis of the Dst index itself. The autocorrelation of the Dst is shown in Figure \ref{fig:dst_correlation}B. Notice the very high autocorrelation of the Dst index with itself for delay times up to $t+7h$. This can also be seen in the lag plot, shown in Figure \ref{fig:dst_correlation}A. This could explain why the persistence model has such high accuracy and correlation when evaluated with the metrics of Section \ref{sect:exp:eval:metrics}, shown in Table \ref{table:LSTMvsPers}. We believe that this also explains why the linear fit parameters of the persistence are so high.

The partial autocorrelation is also an important value. The partial autocorrelation $\alpha(k)$, defined by Equation \eqref{eq:alphaK}, behaves as the autocorrelation between $z_t$ and $z_{t+k}$, adjusted from the intermediate variables $z_{t+1}, z_{t+2}, \dots, z_{t+k-1}$ \citep{box2015time}.
\begin{equation} 
	\alpha(k) =
	\begin{cases}
		\text{cor}(z_{t+1}, z_t) \text{ for } k = 1 \\
		\text{cor}(z_{t+k}-  P_{t,k}(z_{t+k}), z_t - P_{t,k}(z_t)) \text{ for } k \ge 2,
	\end{cases}
	\label{eq:alphaK}
\end{equation}
where $k$ is the lag between the two time series values $z_t$ and $z_{t+k}$, and $P_{t,k}(z)$ is an operator of orthogonal projection of $z$ onto the linear subspace of the Hilbert space spanned by $z_{t+1}, z_{t+2} \dots , z_{t+k}$. The partial autocorrelation of the Dst can be seen in Figure \ref{fig:dst_correlation}C. This shows what can actually be learned from the Dst index: after the correction applied by the autocorrelation, only the Dst at time step $t+2h$ still has some significant correlation to the Dst at time $t$. This would explain why the neural network model has difficulty accurately predicting values beyond $t+1$, and instead relies on behaving as a persistence model to predict the next values.

\hphantom{ }

\subsection{Removing the autocorrelation}
The autocorrelation properties of the Dst index are most likely the causes of the problem in the forecast. Direct workarounds consist of either changing the input or the output. A first solution is not to include the Dst index in the input vectors, as done by \citet{wu1997geomagnetic}. This gives a forecast based purely on the solar wind parameters. 

\begin{figure}
	\centering
	\includegraphics[width=\textwidth]{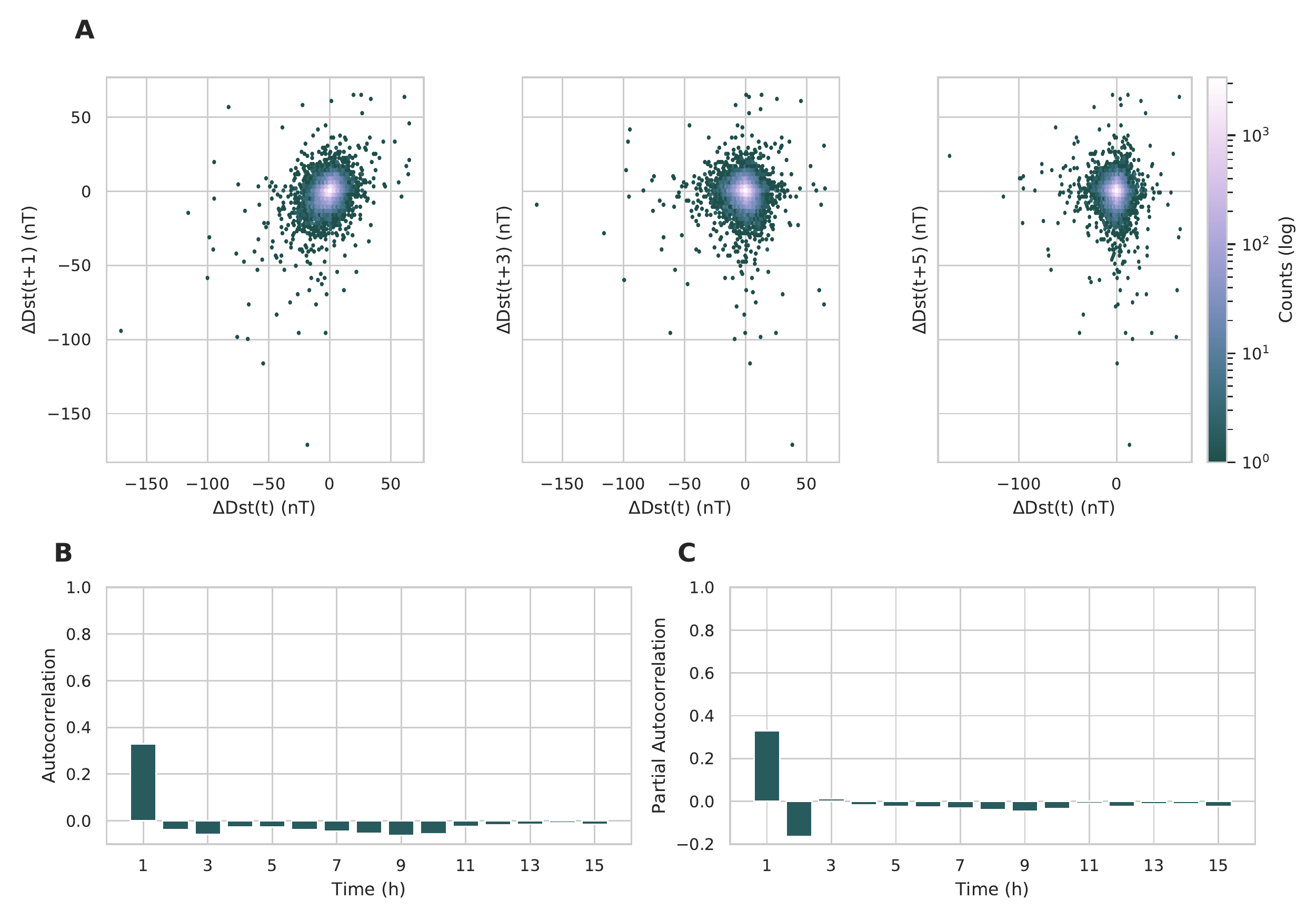}
	\caption{Panel \textbf{(A)} displays a lag plot of the $\Delta$Dst, with lag times of 1, 3 and 5 hours. The color of each point represents the how many times this value was encountered in the data. Notice the lack of any linear relation between the lagged values. Panel \textbf{(B)} displays the autocorrelation of the $\Delta$Dst, panel \textbf{(C)} the partial autocorrelation. Notice that the $\Delta$Dst no longer shows any strong autocorrelation.} \label{fig:delta_dst_correlation}
\end{figure}

Another solution is to de-trend the Dst time series, and instead forecast the change in the Dst. Let us call $\Delta$Dst the difference of the Dst between two time steps:
\begin{equation} 
	\Delta Dst(t) = Dst(t) - Dst(t-1).
	\label{eq:deltadst}
\end{equation}
This parameter has also been introduced by \citet{wintoft2018evaluation}. However, they do not forecast the $\Delta$Dst directly with their model, but use it as a parameter for data selection. A lag plot of $\Delta$Dst shows that the correlation with the previous time step has almost completely vanished, as is visible in Figure \ref{fig:delta_dst_correlation}A. Computing the autocorrelation confirms this, as seen in Figure \ref{fig:delta_dst_correlation}B.

The $\Delta$Dst is a new parameter that we can use to train the LSTM-NN model with. The experiment described in Section \ref{sect:Experiment} is repeated, only this time the model will forecast the $\Delta$Dst model. As input we use, next to the parameters described in Section \ref{sect:exp:data}, also the previous values of the $\Delta$Dst. Table \ref{table:performance_delta_dst} and Table \ref{table:dtw_delta_dst} show the results of the LSTM-NN using this data. The forecasting of the $\Delta$Dst seems to work well for forecasting horizons of 1 to 2 hours. For later forecasting horizons, the correlation coefficient decreases sharply, and the prediction efficiency becomes close to zero. The RMSE does not increase substantially when the forecasting horizon increases. The results of the DTW measure are shown in Table \ref{table:dtw_delta_dst}. Notice the absence of a persistence effect, as most values are no longer on the offside diagonal. Only in the last row does there seem to be some delay, but this could be explained by taking the prediction efficiency of Table \ref{table:performance_delta_dst} into account, which is close to 0 for forecasting horizon $t+6h$. This means that the forecast most likely no longer resembles the observed time series anymore, and the evaluation of the DTW algorithm does not have much meaning anymore.

\begin{table}[!htbp]
	\centering
	\caption {Evaluation of the LSTM-NN model and the persistence model with the metrics from Section \ref{sect:exp:eval:metrics} when forecasting $\Delta$Dst. \label{table:performance_delta_dst}}
	\begin{tabular}{@{}lccccccc@{}}
		\toprule
		Forecasting horizon & RMSE & R & A & B & MAE & ME & PE \\ 
		& (nT) & & & (nT) & (nT) & (nT) & \\ \hline
		\textit{LSTM-NN model} \\
		\quad $t+1h$ & 3.215 & 0.630 & -0.064 & 0.396 & 2.156 & -0.064 & 0.397 \\
		\quad $t+2h$ & 3.807 & 0.393 & -0.031 & 0.163 & 2.505 & -0.029 & 0.154 \\
		\quad $t+3h$ & 3.943 & 0.305 & -0.016 & 0.106 & 2.571 & -0.016 & 0.091 \\
		\quad $t+4h$ & 3.995 & 0.260 & 0.059  & 0.082 & 2.601 & 0.057  & 0.064 \\
		\quad $t+5h$ & 4.057 & 0.197 & -0.018 & 0.053 & 2.640 & -0.021 & 0.033 \\
		\quad $t+6h$ & 4.075 & 0.168 & 0.009  & 0.038 & 2.646 & 0.006  & 0.025 \\
		\textit{Persistence model} \\
		\quad $t+1h$ & 5.160 & 0.311  & 1.729 & 0.311  & 3.524 & 1.730 & -0.553 \\
		\quad $t+2h$ & 6.170 & -0.042 & 1.530 & -0.042 & 4.189 & 1.532 & -1.220 \\
		\quad $t+3h$ & 6.178 & -0.080 & 1.068 & -0.080 & 4.169 & 1.069 & -1.226 \\
		\quad $t+4h$ & 6.078 & -0.063 & 0.685 & -0.063 & 4.065 & 0.686 & -1.155 \\
		\quad $t+5h$ & 6.067 & -0.035 & 1.144 & -0.035 & 4.076 & 1.145 & -1.147 \\
		\quad $t+6h$ & 6.088 & -0.016 & 1.490 & -0.016 & 4.150 & 1.491 & -1.162 \\
		\bottomrule
	\end{tabular}
\end{table}

\begin{table}
	\centering
	\caption{The DTW measure of the LSTM-NN model forecasting the $\Delta$Dst. Notice that the persistence-like behavior is absent.} \label{table:dtw_delta_dst}
	\begin{tabular}{@{}llllllll@{}}
		\toprule
		Forecast horizon & \multicolumn{7}{c}{Time shift} \\  \cmidrule{2-8}
		& 0h    & 1h    & 2h    & 3h    & 4h     & 5h    & 6h \\ \hline
		$t+1h$ & 0.426 & \textbf{0.449} & 0.053 & 0.03  & 0.019 & 0.014 & 0.009 \\
		$t+2h$ & \textbf{0.482} & 0.355 & 0.061 & 0.028 & 0.026 & 0.024 & 0.024 \\
		$t+3h$ & \textbf{0.424} & 0.319 & 0.152 & 0.021 & 0.026 & 0.03  & 0.028 \\
		$t+4h$ & 0.287 & \textbf{0.283} & 0.25  & 0.107 & 0.022 & 0.028 & 0.023 \\
		$t+5h$ & 0.186 & 0.192 & 0.228 & \textbf{0.229} & 0.118 & 0.024 & 0.022 \\
		$t+6h$ & 0.147 & 0.139 & 0.165 & 0.188 & \textbf{0.22}  & 0.127 & 0.013 \\
		\bottomrule
	\end{tabular}
\end{table}

The forecast now accurately shows us that the predictive power of the LSTM is linked to the partial autocorrelation of the $\Delta$Dst, and demonstrates the difficulty to provide an accurate forecast beyond $t+2h$. The advantage of using the $\Delta$Dst is that there is no longer a false sense of accuracy. The persistence effect gave the illusion of a strong forecast, while the $\Delta$Dst does not. Using this as a basis, it will be much more transparent when a forecasting model provides us with an actual accurate forecast.

Finally we discuss the possible causes of why forecasting the Dst is so difficult. We believe that there are two problems that we have not yet taken into account. The first is the variation of the geo-effectiveness of the quiet solar wind, mainly caused by how the tilt of the Earth effects the interaction of magnetosphere with the solar wind. Together with the inclination of the equatorial plane of the Sun, this causes a yearly variation which was not taken into account in this experiment. This is called the Russel-McPherron effect \citep{russell1973semiannual} and has been shown to effect the Dst index \citep{siscoe1996diurnal}. The second is that we believe that it is misguided to forecast the Dst index using the solar wind data measured at L1. These measurements are taken too close to the Earth, which causes an intrinsic limit on how far in the future we can give a forecast. We believe that having measurements at L5 would provide a large improvement in our abilities to provide timely forecasts, as discussed by \citet{hapgood2017l1l5together}. The effects of using measurements at L5 could be explored in future research, where simulations such as EUFHORIA \citep{pomoell2018euhforia} are used to provide artificial measurements.

\section{Conclusions}
\label{sect:conclusion}
An LSTM-based neural network, called the LSTM-NN model, is trained to forecast the Dst index 1 to 6 hours in the future, using solar wind parameters and the Dst from 6 hours before the prediction as an input. While the evaluation scores have indicated that the LSTM-NN model is comparable to the latest publications, visual inspection shows that the model's forecast behavior is similar to that of a persistence model, using the last known input of the Dst as its output. Although the prediction performs better than the persistence model, showing that some information can be learned from the solar wind, the LSTM-NN model effectively fails in its objective. 

In order to detect this new type of error, a new test is developed based on the DTW algorithm, to measure the shift between observation and model prediction. DTW can compare two time series in a time window, instead of comparing two values on the same timestamp such as done by the RMSE and the correlation coefficient, allowing the detection of temporal trends. By using the output of the DTW algorithm, first a least-distance mapping is given between the two time series, which can then be used to compare the timestamps of the points mapped to each other. This gives us a measure of the time warp between these two time series, from which we can infer a potential persistence effect. 
    
When this new measure was applied to the persistence model, the results were as expected, and completely captured the temporal behavior of the persistence model. When the measure was applied to the time series forecasting of the LSTM-NN model, it detected the temporal lag in the forecast, proving its usefulness. 

Finally, the possible origin of this lag was discussed by observing the autocorrelation of the time series, together with possible different experiments that do not suffer this temporal lag. It was shown that the forecasting of the differentiated Dst did not have this temporal lag. The LSTM-NN model showed promising results for forecasting horizons of t+1h and t+2h, but later forecasts did not have a very high accuracy or correlation to the observations. Future studies focusing on forecasting the differentiated Dst could provide more transparent results. We believe that new research also has to explore the effect of the variability of the solar wind interacting with the magnetosphere in function of the Earth tilt and the inclination of the solar equatorial plane. 

Finally, we believe that the observational data measured at L1 plays a big role in limiting the forecast horizon of the Dst index. Looking at the effects of having measurements at L5 should be further explored in future work, using simulations to provide the artificial measurements.

As a concluding remark, we would like to emphasize that researchers should be very prudent when reporting results of time series forecasting with the metrics defined in Section \ref{sect:exp:eval:metrics}. These metrics fail to capture behaviors that are only seen when taking into account the temporal dimension of the forecasting, and could provide misleading results.

\begin{appendices}
\section{Additional information}
\label{appendix:additionalInfo}
Training and evaluating the model takes on average 10 minutes on a machine with the following specifications:
\begin{itemize}
	\item \textbf{OS:} Windows 10
	\item \textbf{Processor:} Intel(R) Core(TM) i7-8850H CPU @ 2.60GHz, 2592 Mhz, 6 Core(s), 12 Logical Processor(s)
	\item \textbf{RAM:} 16Gb 
\end{itemize}	

The source code and experiment can be found on the following webpage: \url{https://github.com/brechtlaperre/DTW_measure}.

\end{appendices}

\section*{Conflict of Interest Statement}

The authors declare that the research was conducted in the absence of any commercial or financial relationships that could be construed as a potential conflict of interest.

\section*{Author Contributions}
BL performed and analyzed the experiment, developed and tested the new technique and wrote the manuscript. JA planned the study and has provided substantial intellectual contribution and interpretation of the results. All authors took part in the manuscript revision and have read and approved the submitted version.

\section*{Acknowledgment}

This research has received funding from the European Union’s Horizon 2020 research and innovation program under grant agreement No 776262 (AIDA, \url{www.aida-space.eu}). 
The OMNI data used for this research were obtained from the GSFC/SPDF OMNIWeb interface at \url{https://omniweb.gsfc.nasa.gov}.
The code will be made publicly available after publication on the project's website at \url{www.aida-space.eu}.


\bibliographystyle{abbrv}
\bibliography{bibliography}

\end{document}